\documentclass[10pt,twocolumn,letterpaper]{article}

\usepackage{iccv}
\usepackage{times}
\usepackage{epsfig}
\usepackage{graphicx}
\usepackage{amsmath}
\usepackage{amssymb}
\usepackage{authblk}

\usepackage{caption}
\usepackage{subcaption}

\usepackage{array}
\usepackage{tabulary}
\newcolumntype{K}[1]{>{\centering\arraybackslash}p{#1}}

\usepackage{algpseudocode,algorithm,algorithmicx}
\newcommand*\Let[2]{\State #1 $\gets$ #2}

\algrenewcommand\algorithmicrequire{\textbf{Input:}}
\algrenewcommand\algorithmicensure{\textbf{Output:}}

\usepackage[pagebackref=true,breaklinks=true,letterpaper=true,colorlinks,bookmarks=false]{hyperref}

\iccvfinalcopy 

\def\iccvPaperID{808} 

\ificcvfinal\fi

\begin{document}

	\title{Deep Reinforcement Learning for Visual Object Tracking in Videos}
	\author[1]{Da Zhang}
    \author[2]{Hamid Maei}
    \author[1]{Xin Wang}
    \author[1]{Yuan-Fang Wang}
    \affil[1]{Department of Computer Science, University of California at Santa Barbara}
    \affil[2]{Samsung Research America}
    \affil[ ]{\tt\small \{dazhang, xwang, yfwang\}@cs.ucsb.edu hamid.maei@samsung.com}

\maketitle

\begin{abstract}
	In this paper we introduce a fully end-to-end approach for visual tracking in videos that learns to  predict the bounding box locations of a target object at every frame. An important insight is that the tracking problem can be considered as a sequential decision-making process and historical semantics encode highly relevant information for future decisions. Based on this intuition, we formulate our model as a recurrent convolutional neural network agent that interacts with a video overtime, and our model can be trained with reinforcement learning (RL) algorithms to learn good tracking policies that pay attention to continuous, inter-frame correlation and maximize tracking performance in the long run. The proposed tracking algorithm achieves state-of-the-art performance in an existing tracking benchmark and operates at frame-rates faster than real-time. To the best of our knowledge, our tracker is the first neural-network tracker that combines convolutional and recurrent networks with RL algorithms.
\end{abstract}

\section{Introduction}

Given some object of interest marked in one frame of a video, the goal of \emph{single-object tracking} is to locate this object in subsequent video frames, despite object movement, changes in the camera's viewpoint and other incidental environmental variations such as lighting and shadows. Single-object tracking finds immediate applications in many important scenarios such as autonomous driving, unmanned aerial vehicle, security surveillance, etc.

Despite the success of traditional trackers based on low-level, hand-crafted features~\cite{babenko2011robust,hare2011struck,ross2008incremental}; models based on deep convolutional neural network (CNN) have dominated recent visual tracking research~\cite{nam2015learning,held2016learning,bertinetto2016fully}. The success of all these models largely depends on the capability of CNN to learn a good feature representation for the tracking target. In order to predict the target location in a new frame, either a search-and-classify~\cite{nam2015learning} or crop-and-regress~\cite{held2016learning,bertinetto2016fully} approach is applied. In that sense, although the representation power of CNN is exploited to capture \emph{spatial} features, only limited manual \emph{temporal} constraints are added in these frameworks, \eg, new target lies in the spatial vicinity of the previous prediction.  Unfortunately, for a busy scene with multiple occluding objects, short-term cues of correlating temporally close objects can often fail to account for multiple targets and mutual occlusion. Hence,  how to harness the power of deep-learning models to automatically learn both \emph{spatial} and \emph{temporal} constraints, especially with longer-term information aggregation and disambiguation, should be fully explored. 

\begin{figure*}
\begin{center}
\includegraphics[width=0.95\linewidth]{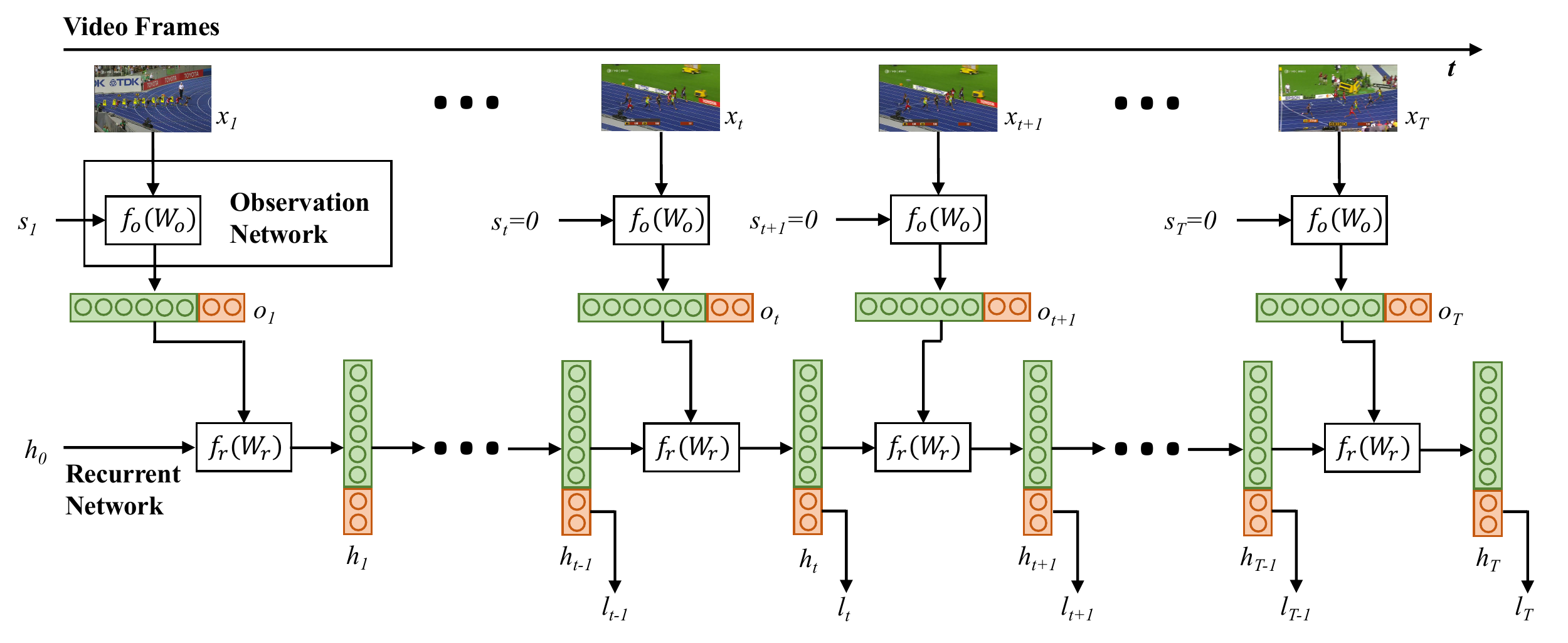}
\end{center}
   \caption{Overview of our Deep RL Tracker: At each timestep $t$, the \textbf{observation network} takes an image $x_t$ and a location vector $s_t$ as input and computes a feature representation $o_t$, where $s_1$ is the ground-truth location at the first frame and $s_t=0$ otherwise. The \textbf{recurrent network} takes $o_t$ as input, combining with the previous hidden state $h_{t-1}$, and generates new hidden state $h_t$. The predicted target location is directly extracted from $h_t$. During training, the agent will receive a reward signal $r_t$ for each prediction. The basic RNN iteration is repeated for a variable number of steps and the cumulative rewards $R=\sum_{t=1}^{T}r_t$ will be used to update network parameters such that the long-term tracking performance is maximized.}
\label{fig:overview}
\end{figure*}

Inspired by the successful works that have applied recurrent neural network (RNN) on computer vision tasks such as video classification and visual attention~\cite{donahue2015long,mnih2014recurrent}. We explore and investigate a more general strategy to develop a novel visual tracking approach based on recurrent convolutional networks. The major intuition behind our method is that the historical visual semantics and tracking proposals encode pertinent information for future predictions and can be modeled as a recurrent convolutional network. However, unlike video classification or visual attention where only high-level semantic or single-step predictions are needed, visual tracking requires continuous and accurate predictions in both spatial and temporal domain over a long period of time, and thus, requires a novel network architecture design as well as proper training algorithms.

In this work, we formulate the visual tracking problem as a sequential decision-making process and propose a novel framework, referred to as \emph{Deep RL Tracker (DRLT)}, which processes video frames as a whole and directly outputs location predictions of the target in each frame. Our model integrates convolutional network with recurrent network (Figure~\ref{fig:overview}), and builds up a spatial-temporal representation of the video. It fuses past recurrent states with current visual features to make predictions of the target object's location over time. We describe an end-to-end RL algorithm that allows the model to be trained to maximize tracking performance in the long run. This procedure uses backpropagation to train the nueral-network components and REINFORCE algorithm~\cite{williams1992simple} to train the policy network.

Our algorithm augments traditional CNN with a recurrent convolutional model learning \emph{spatial-temporal} representations and RL to maximize long-term tracking performance. The main contributions of our work are:

\begin{itemize}
    \item We propose and develop a novel convolutional recurrent neural network model for visual tracking. The proposed method directly leverages the power of deep-learning models to automatically learn both spatial and temporal constraints. 
    \item Our framework is trained end-to-end with deep RL algorithms, in which the model is optimized to maximize a tracking performance measure in the long run. 
    \item Our model is trained fully off-line. When applied to online tracking, only a single forward pass is computed and no online fine-tuning is needed, allowing us to run at frame-rates beyond real-time.
    \item Our extensive experiments demonstrate the outstanding performance of our tracking algorithm compared to the state-of-the-art techniques in OTB~\cite{wu2013online} public tracking benchmark.
\end{itemize}

We claim that recurrent convolutional network plus RL algorithm is another useful deep-learning framework apart from CNN-based trackers. It has the potential of developing into a much robust and accurate tracker given that it pays explicit attention to temporal correlation and a long-term reward mechanism through RL.

The rest of the paper is organized as follows. We first review related work in Section~\ref{sec:relatedwork}, and discuss our RL approach for visual tracking in Section~\ref{subsec:arch}. Section~\ref{subsec:training} describes our end-to-end optimization algorithm, and Section~\ref{sec:exp} demonstrates the experimental results using a standard tracking benchmark. 

\section{Related Work}
\label{sec:relatedwork}

\subsection{Visual Object Tracking}

Visual Tracking is a fundamental problem in computer vision that has been actively studied for decades. Many methods have been proposed for single-object tracking. For a systematic review and comparison, we refer the readers to a recent benchmark and a tracking challenge report~\cite{wu2013online, kristan2015visual}.

\textbf{Classification-based trackers.} Trackers for generic object tracking often follows a tracking-by-classification methodology~\cite{kalal2012tracking,wang2015understanding}. A tracker will sample "foreground" patches near the target object and "background" patches farther away from the target. These patches are then used to train a foreground-background classifier, and this classifier is used to score potential patches in the next frame to estimate the new target location. Usually, the classifier is first trained off-line and fine-tuned during online tracking. Many neural-network trackers following this approach~\cite{hong2015online,nam2015learning,wang2015visual,zhang2015robust} have surpassed traditional trackers~\cite{babenko2011robust,hare2011struck,ross2008incremental}, and achieved state-of-the-art performance~\cite{nam2015learning,kristan2015visual}. Unfortunately, these trackers are inefficient at run-time since neural networks are very slow to train in an online fashion. Another drawback of such a design is that it does not fully utilize all video information, particularly explicit temporal correlation. 

\textbf{Regression-based trackers.} Some recent works~\cite{held2016learning,bertinetto2016fully} have attempted to treat tracking as a regression instead of classification problem. David \emph{et al.}~\cite{held2016learning} trained a CNN to regress directly from two images to the location in the second image of the object shown in the first image. Luca \emph{et al.}~\cite{bertinetto2016fully} proposed a fully-convolutional siamese network to track objects in videos. These deep-learning methods can run at frame-rates beyond real time while maintaining state-of-the-art performance. However, they only extract features independently  from each video frame and only perform comparison between two consecutive frames, prohibiting them  from fully utilizing longer-term contextual and temporal information.

\textbf{Recurrent-neural-network trackers.} Several recent works~\cite{kahou2015ratm,gan2015first} have sought to train recurrent neural networks for the problem of visual tracking. Gan~\emph{et al.}~\cite{gan2015first} trained an RNN to predict the absolute position of the target in each frame and Kahou \emph{et al.}~\cite{kahou2015ratm} similarly trained an RNN for tracking using the attention mechanism. Although they brought good intuitions from RNN, these methods have not yet demonstrated competitive results on modern benchmark. 

Another related work to ours is~\cite{ning2016spatially}. They proposed a spatially supervised recurrent convolutional neural network in which a YOLO network~\cite{redmon2016you} is applied on each frame to produce object detections and a recurrent neural network is used to directly regress YOLO detections. Our framework does not need any supervision from other detection module and is more general and flexible.

\subsection{Deep Reinforcement Learning}

RL is a learning method based on trial and error, where an agent does not necessarily have {\em a prior} knowledge about what is the correct action to take. It learns interactively from rewards fed back from the environments. In order to maximize the expected rewards in the long term, the agent learns the best policy.

We draw inspiration from recent approaches that have used REINFORCE~\cite{williams1992simple} to learn task-specific policies. Mnih \emph{et al.}~\cite{mnih2014recurrent} and Ba \emph{et al.}~\cite{ba2014multiple} learned spatial attention policies for image classification, and Xu \emph{et al.}~\cite{xu2015show} for image caption generation. Our work is similar to the attention model described in~\cite{mnih2014recurrent}, but we designed our own network architecture specially tailored for solving the visual tracking problem by combining CNN, RNN and RL algorithms.

Our proposed framework directly apply RNN on top of frame-level CNN features, paying direct attention to both \emph{spatial} and \emph{temporal} constraints, and the full framework is trained off-line with REINFORCE algorithm in an end-to-end manner. Due to its run-time simplicity, our tracker runs at frame-rates beyond real-time while maintaining state-of-the-art performance. We will describe our framework in detail in Section~\ref{sec:DRLT}.

\section{Deep RL Tracker (DRLT)}
\label{sec:DRLT}

Our goal is to take a sequence of video frames and output target object locations at each frame. We formulate our tracking algorithm as a sequential decision-making process of a goal-oriented agent interacting with the visual environment. Figure~\ref{fig:overview} shows our model structure. At each point in time, the agent extracts representative features from a video frame, integrates information over time, and decides how to take actions accordingly. The agent receives a scalar reward signal at each timestep, and the goal of the agent is to maximize the total long-term rewards. Hence, it must learn to effectively utilize these temporal observations to reason on the moving trajectory of the object.

\subsection{Architecture}
\label{subsec:arch}

The model consists of two major components: an observation network (Section~\ref{subsec:obsernet}), and a recurrent network (Section~\ref{subsec:recurnet}). The \emph{observation network} encodes representations of video frames. The \emph{recurrent network} integrates these observations over time and predicts the bounding box location in each frame. We now describe each of these in more detail. Later in Section~\ref{subsec:training}, we explain how we use a combination of backpropagation and REINFORCE to train the model in an end-to-end fashion. 
 
\subsubsection{Observation Network}
\label{subsec:obsernet}

As shown in Figure~\ref{fig:overview}, the observation network $f_o$, parameterized by $W_o$, observes a single video frame $x_t$ at each timestep. It encodes the frame into a feature vector $i_t$, concatenates a location vector $s_t$ and provides the feature and location combo (denoted as $o_t$) as input to the recurrent network.

The feature vector $i_t$ is typically computed with a sequence of convolutional, pooling, and fully connected layers to encode information about \emph{what} was seen in this frame. The importance of $s_t$ are two folds: When the ground-truth bounding box location is known, such as the first frame in a given sequence, $s_t$ is directly set to be the normalized location coordinate $(x,y,w,h)\in [0,1]$, serving as a strong supervising guide for further inferences. Otherwise, $s_t$ is padded with zero and only the feature information $i_t$ is incorporated by  the recurrent network.

The concatenation of $i_t$ and $s_t$ allows the recurrent network to directly encode image features as well as location predictions, and it is also easier for location regression.

\subsubsection{Recurrent Network}
\label{subsec:recurnet}

The recurrent network $f_r$, parameterized by $W_r$, forms the core of the learning agent. As can be seen in Figure~\ref{fig:overview}, at one single timestep $t$, the observation feature vector $o_t$ is fed into a recurrent network, and the recurrent network updates its internal hidden state $h_t$ based on the previous hidden state $h_{t-1}$ and the current observation feature vector $o_t$: 
\begin{equation}
	h_t=f_{r}(h_{t-1},o_t;W_r)
    \label{eq:recurnet}
\end{equation}
where $f_r$ is a recurrent transformation function and we use LSTM~\cite{hochreiter1997long} in our network.

Importantly, the network's hidden state $h_t$ models temporal hypotheses about target object locations. Since $o_t$ is a concatenation of the image feature and the location signal, $h_t$ directly encodes information about both \emph{where} in the frame an object was located as well as $what$ was seen. 

As the agent reasons on a video, it outputs the location of target object $l_t$ at each timestep $t$. $l_t=(x,y,w,h)$
where $(x,y)$ represent the coordinates of the bounding box center relative to the width and height of the image, respectively. The width and height of the bounding box are also relative to those of the image, consequently, $(x,y,w,h)\in [0,1]$.

The predicted location $l_t$ is directly extracted from the last four elements of $h_t$ denoted as $\mu_t$, such that the agent's decision is a function of its past observations and their predicted locations. At training time, $l_t$ is sampled from a multi-variate Gaussian distribution with a mean of $\mu_t$ and a fixed variance; at test time, the maximum a posteriori estimate is used. 

Figure~\ref{fig:overview} further illustrates the roles of each component as well as the corresponding inputs and outputs with an example of a forward pass through the network.
\subsection{Training}
\label{subsec:training}

Training this network to maximize the overall tracking performance is a non-trivial task, and we leverage the REINFORCE algorithm~\cite{williams1992simple} from the RL community to solve this problem.

\subsubsection{Reward Function}

During training, the agent will receive a reward signal $r_t$ from the environment after executing an action at time $t$. In this work, we explore two different reward definitions in different training phases. One is 
\begin{equation}
	r_t=-avg(|l_t-g_t|)-max(|l_t-g_t|)
    \label{eq:rewardefinition_early}
\end{equation}
where $l_t$ is the location outputted by the recurrent network, $g_t$ is the target ground truth at time $t$, $avg(\cdot)$ and $max(\cdot)$ compute the pixel-wise mean and maximum. The other reward is 
\begin{equation}
	r_t=\frac{|l_t\cap g_t|}{|l_t\cup g_t|}
    \label{eq:rewardefinition_late}
\end{equation}
where the reward is computed as the intersection area divided by the union area (IoU) between $l_t$ and $g_t$. 

The training objective is to maximize the sum of the reward signals: $R=\sum_{t=1}^{T}r_t$. By definition, the reward in Equation \ref{eq:rewardefinition_early} and Equation \ref{eq:rewardefinition_late} both measure the closeness between predicted location $l_t$ and ground-truth location $g_t$. We use the reward definition in Equation \ref{eq:rewardefinition_early} in the early stage of training, while using the reward definition in Equation \ref{eq:rewardefinition_late} in the late stage of training to directly maximize the IoU  between the prediction $l_t$ and ground-truth $g_t$.

\subsubsection{Gradient Approximation}
	Our network is parameterized by $W=\{W_o,W_r\}$ and we aim to learn these parameters to maximize the total tracking reward the agent can expect in the long run. More specifically, the objective of the agent is to learn a policy function $\pi(l_t|z_{1:t};W)$ with parameters W that, at each step $t$, maps the history of past interactions with the environment $z_{1:t}=x_1,l_1,x_2,l_2,...,x_{t-1},l_{t-1},x_t$ (a sequence of past observations and actions taken by the agent) to a distribution over actions for the current timestep. Here, the policy $\pi$ is defined by our neural network architecture, and the history of interactions $z_{1:t}$ is summarized in the hidden state $h_t$. For simplicity, we will use $Z_t=z_{1:t}$ to indicate all histories up to time $t$, thus, the policy function can be written as $\pi(l_t|Z_t;W)$.
    
    To put it in a formal way, the policy of the agent $\pi(l_t|Z_t;W)$ induces a distribution over possible interactions $Z_t$ and we aim to maximize the total reward $R$ under this distribution, thus, the objective is defined as:
\begin{equation}
	G(W)=E_{p(z_{1:T};W)}[\sum_{t=1}^{T}r_t]=E_{p(Z_T;W)}[R]
    \label{eq:rlagentori}
\end{equation}
where $p(z_{1:T};W)$ is the distribution over possible interactions parameterized by $W$.
    
    This formulation involves an expectation over high-dimensional interactions which is hard to solve in traditional supervised manner. Here, we bring techniques from the RL community to solve this problem, as shown in~\cite{williams1992simple}, the gradient can be first simplified by taking the derivative over log-probability of the policy function $\pi$:
\begin{equation}
	\nabla_{W}G=\sum_{t=1}^{T}E_{p(Z_T;W)}[\nabla_{W}\ln\pi(l_t|Z_t;W)R]
    \label{eq:gradientfirst}
\end{equation}
and the expectation can be further approximated by an episodic algorithm: since the action is drawn from probabilistic distributions, one can execute the same policy for many episodes and approximate expectation by taking the average, thus
\begin{equation}
	\nabla_{W}G\approx\frac{1}{N}\sum_{i=1}^{N}\sum_{t=1}^{T}\nabla_{W}\ln\pi(l_t^i|Z_t^i;W)R^i
    \label{eq:gradientsecond}
\end{equation}
where $R^i$s are cumulative rewards obtained by running the current policy $\pi$ for $N$ episodes, $i=1...N$. 
    
    The above training rule is known as the episodic REINFORCE~\cite{williams1992simple} algorithm, and it involves running the agent with its current policy to obtain samples of interactions and then updating parameters $W$ of the agent such that the log-probability of  chosen actions that have led to high overall rewards is increased.
    
    In practice, although Equation \ref{eq:gradientsecond} computes a good estimation of the gradient $\nabla_{W}G$, when applied to train the deep RL tracker, the training process is hard to converge due to the high variance of this gradient estimation. Thus, in order to obtain an unbiased low-variance gradient estimation, a common method is to subtract a reinforcement baseline from the cumulative rewards $R$:
\begin{equation}
	\nabla_{W}G\approx\frac{1}{N}\sum_{i=1}^{N}\sum_{t=1}^{T}\nabla_{W}\ln\pi(l_t^i|Z_t^i;W)(R_t^i-b_t)
    \label{eq:gradientfinal}
\end{equation}
where $b_t$ is called reinforcement baseline in the RL literature, it is natural to select $b_t=E_{\pi}[R_t]$, and this form of baseline is known as the value function~\cite{sutton1999policy}. This estimation maintains the same expectation with Equation \ref{eq:gradientsecond} while sufficiently reduces the variance. 
    
    \subsubsection{Training with Backpropagation}
    
    The only remaining part to compute the gradient in Equation \ref{eq:gradientfinal} is to compute the gradient over log-probability of the policy function $\nabla_{W}\ln\pi(l_t|Z_t;W)$. To simplify notation, we focus on one single timestep and omit usual unit index subscript throughout. In our network design, the policy function $\pi$ outputs the target location $l$ which is drawn from a Gaussian distribution centered at $\mu$ with fixed variance $\sigma$, and $\mu$ is the output of the deep RL tracker parameterized by $W$. The density function $g$ determining the output $l$ on any single trial is given by:
\begin{equation}
	g(l,\mu,\delta)=\frac{1}{(2\pi)^{\frac{1}{2}}\sigma}e^{-\frac{(l-\mu)^2}{2\sigma^2}}
    \label{eq:densityfunction}
\end{equation}    
    
    Based on REINFORCE algorithm~\cite{williams1992simple}, the gradient of the policy function with respect to $\mu$ is given by the gradient of the density function:
\begin{equation}
	\nabla_{\mu}\ln\pi=\frac{\partial \ln g}{\partial \mu}=\frac{l-\mu}{\sigma^2}
    \label{eq:gradientgaussian}
\end{equation}
since $\mu$ is the output of deep RL tracker parameterized by $W$, the gradients with respect to network weights $W$ can be easily computed by standard backpropagation.

	\subsubsection{Overall Procedure}
    
    The overall procedure of our training algorithm is presented in Algorithm \ref{alg:rltraining}. The network parameters $W$ are first randomly initialized to define our initial policy. Then, we take first $T$ frames from one training video to be the input of our network. We execute current policy $N$ times, compute gradients and update network parameters. Next, we take consecutive $T$ frames from the same video and apply the same training procedure. We repeat this for all training videos in our dataset, and we stop when we reach the maximum number of epochs or the cumulative reward ceases to increase.
    
    \begin{algorithm}
      \caption{Deep RL Tracker training algorithm}
        \label{alg:rltraining}
      \begin{algorithmic}[1]
        \Require{Training videos $\{v_1,...,v_M\}$} with ground-truth
        \Ensure{Network weights $W$}
        \State Randomly initialize weights $W_o$ and $W_r$
        \State Start from the first frame in training dataset
        \Repeat
            \State Sequentially select $T$ frames $\{x_1,...,x_T\}$
            \State Extract features $\{o_1,...,o_T\}$
            \State Generate hidden states $\{h_1,...,h_T\}$
            \State Compute network output $\{\mu_1,...,\mu_T\}$
            \State Randomly sample predictions for $N$ episodes $\{l_1^{1:N},...,l_T^{1:N}\}$ according to Equation \ref{eq:densityfunction}
            \State Calculates rewards $\{r_1^{1:N},...,r_T^{1:N}\}$ based on Equation \ref{eq:rewardefinition_early} in early iterations or Equation \ref{eq:rewardefinition_late} in late iterations
            \Let{$b_t$}{$\frac{1}{N}\sum_{i=1}^{N}r_t^i, t=1,...,T$}
            \State Computes gradient according to Equation \ref{eq:gradientfinal}
            \State Update $W$ using backpropagation
            \State Move on to next $T$ frames
        \Until{reward doesn't increase}
      \end{algorithmic}
    \end{algorithm}
    
    During testing, the network parameters $W$ are fixed and no online fine-tuning is needed. The procedure at test time is as simple as computing one forward pass of our algorithm, \emph{i.e.,} given a test video, the deep RL tracker predicts the location of target object in every single frame by sequentially processing the video data. 
\section{Experimental Results}
\label{sec:exp}

We evaluated the proposed approach of visual object tracking on the Object Tracking Benchmark~\cite{wu2015object}, and compared its performance with state-of-the-art trackers. Our algorithm was implemented in Python using TensorFlow toolbox\footnote {https://www.tensorflow.org/}, and ran at around 45 fps with an NVIDIA GTX 1080 GPU.

\subsection{Evaluation Metrics}
We followed the evaluation protocols in~\cite{wu2013online}, where the performance of trackers was measured based on two different metrics: success rate and precision plots. In both metrics, the ratio of successfully tracked frames was measured by a set of thresholds, where bounding box overlap ratio and center location error were employed in success rate plot and precision plot, respectively. We ranked the tracking algorithms based on the Area-Under-Curve (AUC) for the success rate plot and center location error at 20 pixels for the precision plot, again, following~\cite{wu2013online}. We also compared the average bounding box overlap ratio for each tracking sequence, as well as run-time tracking speed.

\begin{figure}[t]
\begin{center}
   \includegraphics[width=0.95\linewidth]{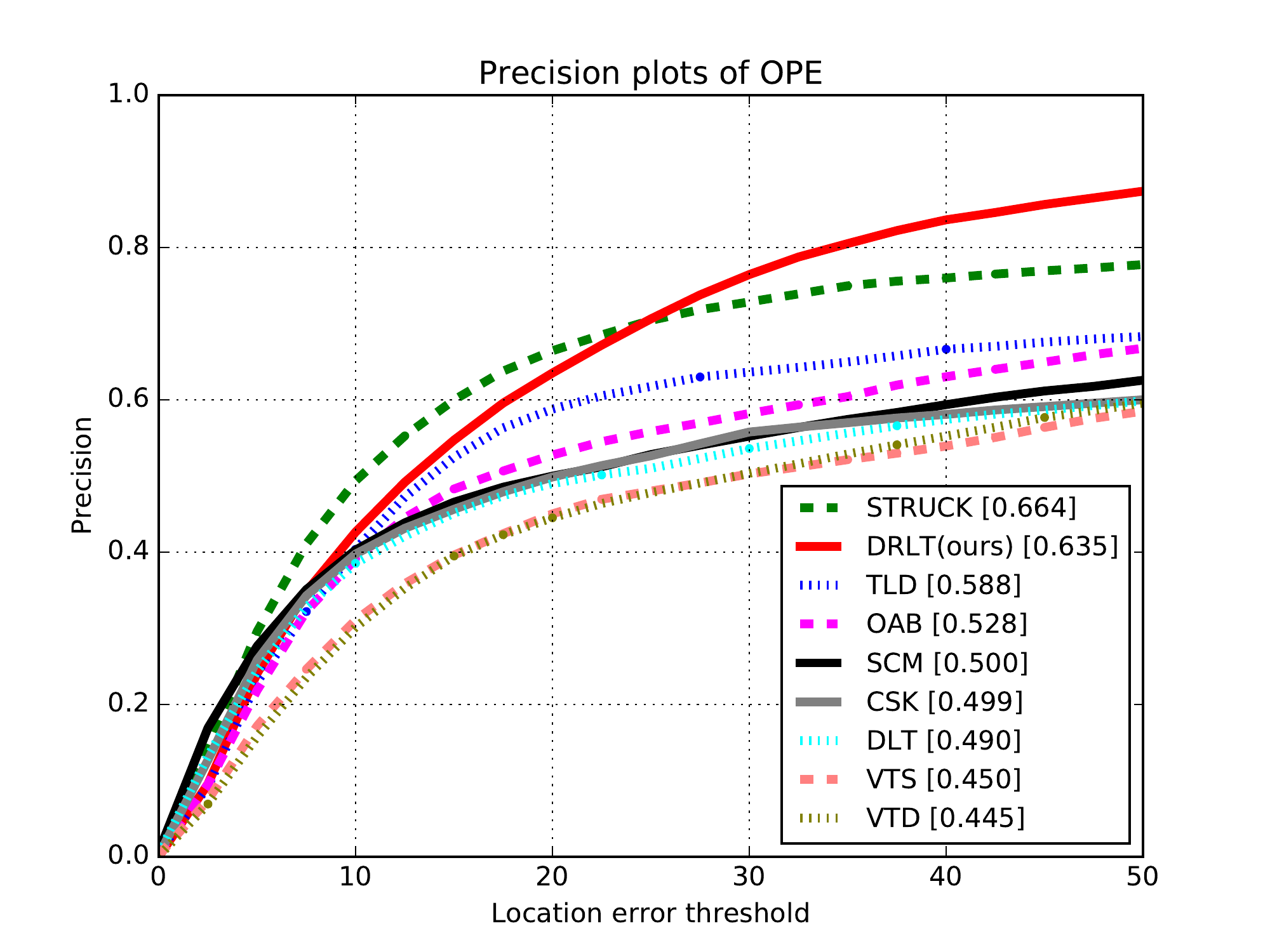}
   
   \includegraphics[width=0.95\linewidth]{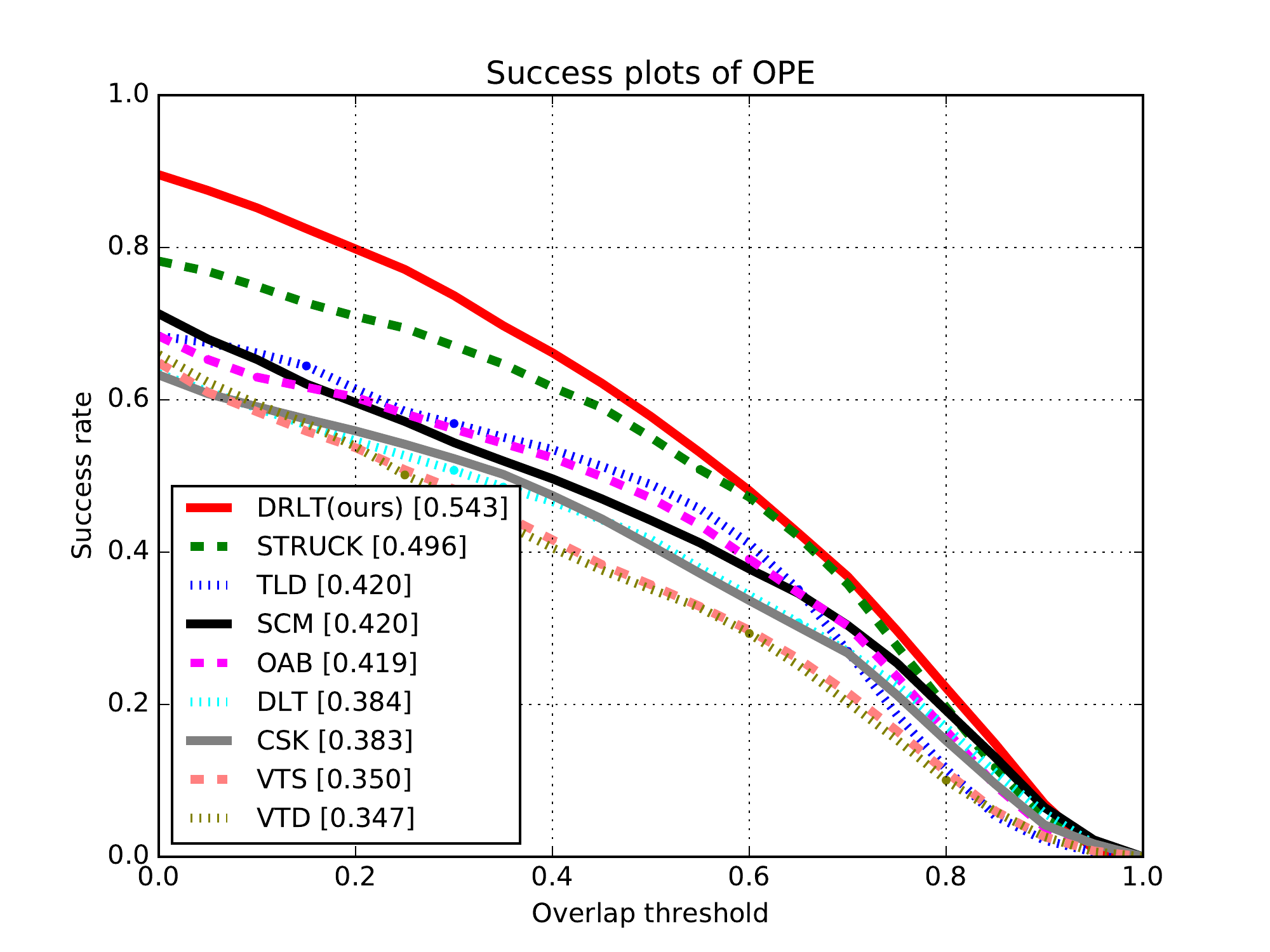}
\end{center}
   \caption{Precision and success plots on a subset of benchmark. The numbers in the legend indicate the representative precisions at 20 pixels for precision plot, and the AUC scores for success plots.}
\label{fig:OPE_benchmark}
\end{figure}

\subsection{Implementation Details}
Here, we describe the design choices of our \emph{observation network} and \emph{recurrent network} as well as the \emph{network learning} procedure in detail.

\textbf{Observation network:} We used a YOLO network~\cite{redmon2016you} fine-tuned on the PASCAL VOC dataset~\cite{pascal-voc-2012} to extract visual features from observed video frames as YOLO was both accurate and time-efficient. The first Fc-layer features were extracted and concatenated with the location vector into a 5000-dimensional vector. Since the pre-trained YOLO weights were fixed during training, we added one more Fc-layer, with 5000 neurons on top of the concatenated vector, and provided the final observation vector as the input to the recurrent network.

\textbf{Recurrent network:} We used a 1-layer LSTM network with 5000 hidden units. At each timestep $t$, the last 4 digits were directly taken as the mean value $\mu$ of the location policy $l$. The location policy was sampled from a Gaussian distribution with mean $\mu$ and variance $\sigma$ during training, and we found that $\sigma=10^{-2}$ was good for both randomness and certainty in our experiment. During testing, we directly used the output mean value $\mu$ as prediction which was the same as setting $\sigma=0$. 

\textbf{Network learning:} The training algorithm was the same as Algorithm \ref{alg:rltraining}, we used $T=10$ and $N=5$ as these hyper-parameters provided the best tracking performance. We kept the pre-trained YOLO weights unchanged as they were proven to encode good information for both semantic prediction and localization, while the weights of the Fc-layer and the LSTM were updated using ADAM algorithm~\cite{kingma2014adam}. The initial learning rate was $10^{-5}$ and we exponentially annealed the learning rate from its initial value to $10^{-6}$ over the course of training. We trained the model up to 500 epochs, or until the cumulative tracking reward $R$ stopped increasing. The first 300 epochs were trained with reward defined in Equation \ref{eq:rewardefinition_early} while the last 200 epochs were trained with reward defined in Equation \ref{eq:rewardefinition_late}.

The trained model was directly applied to the test sequences with no online fine-tuning. During testing, only the video frames and the ground-truth location in the first frame were inputed to the network, and the predictions were directly generated through a single forward pass of the recurrent network.

\begin{figure*}
\begin{center}
\includegraphics[width=\linewidth]{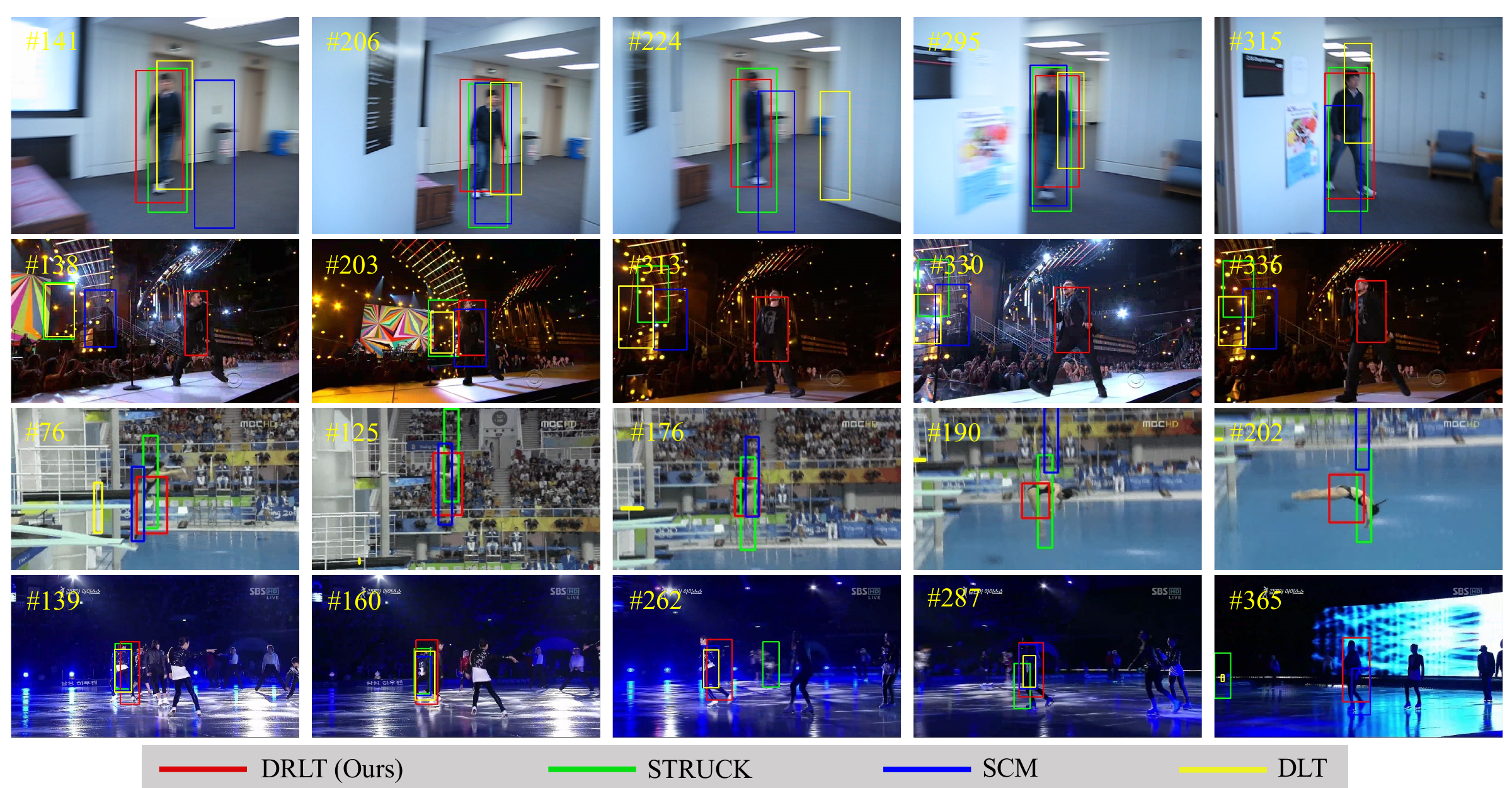}
\end{center}
   \caption{Qualitative results of the proposed method on some challenging sequences (\emph{BlurBody}, \emph{Singer2}, \emph{Diving}, \emph{Skating1}).}
\label{fig:qualitative}
\end{figure*}

\subsection{Evaluation on benchmark}

We have conducted extensive experiments on comparing the performance of our algorithm with eight other distinct trackers on a suite of 30 challenging and publicly available video sequences. Specifically, the one-pass evaluation (OPE)~\cite{wu2013online} was employed to compare our algorithm with seven top trackers included in the benchmark suite: STRUCK~\cite{hare2011struck}, TLD~\cite{kalal2012tracking}, CSK~\cite{henriques2012exploiting}, OAB~\cite{grabner2006real}, VTD~\cite{kwon2010visual}, VTS~\cite{kwon2011tracking}, SCM~\cite{zhong2012robust}. Note that DLT~\cite{wang2013learning} was another tracking algorithm based on deep neural networks, which provided a baseline for tracking algorithms adopting deep learning. Since the YOLO weights were pre-trained on ImageNet dataset and finetuned on PASCAL VOC, capable of detecting objects of 20 classes, we picked a subset of 30 videos from the benchmark where the targets belonged to these classes (Table~\ref{tb:seq_comp_table}). According to our evaluation results, the difficulty of this subset was harder than that of the full benchmark.

\begin{figure}[t]
\begin{center}
   \includegraphics[width=0.95\linewidth]{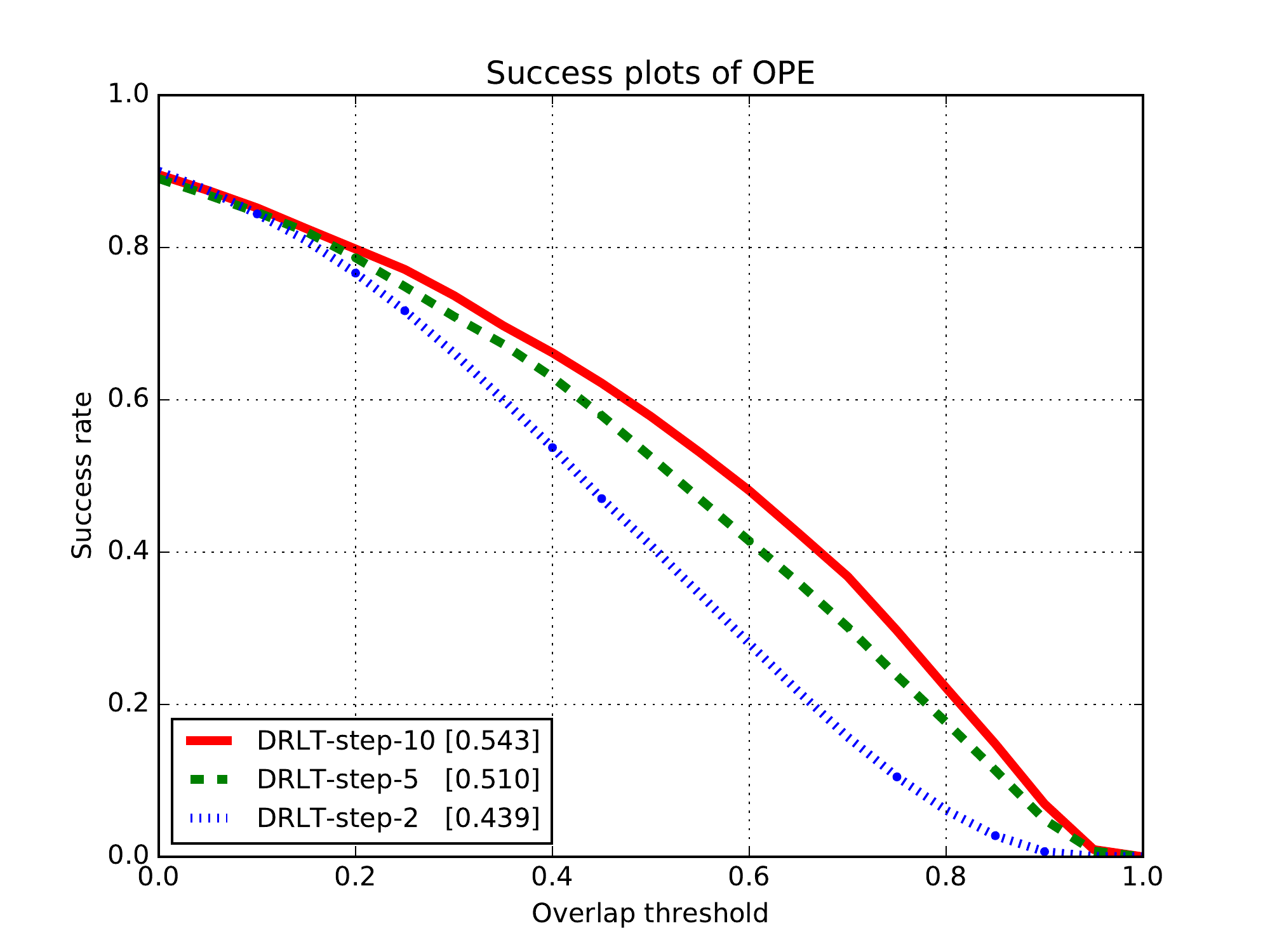}
\end{center}
   \caption{Success plots on a subset of benchmark for varying RNN step sizes. The numbers in the legend indicate the AUC scores.}
\label{fig:internal}
\vspace{-3ex}
\end{figure}

As a generic object tracker, our algorithm was trained off-line and no online fine tuning mechanisms were applied. Thus, training data with similar dynamics were needed to capture both categorical and motional information. We split the dataset and used first $1/3$ frames in each sequence with ground truth for off-line training, while the algorithm was tested on the whole sequence with unseen frames. This property made our algorithm especially useful in surveillance environments, where models could be trained off-line with pre-captured data.

\begin{table}[t]
\footnotesize
\begin{center}
\begin{tabular}{|c|c|c|c|}
\hline
Tracker & AUC & precision & speed (fps)\\
\hline\hline
DLT~\cite{wang2013learning} & 0.384 & 0.490 & 8 \\
STRUCK~\cite{hare2011struck} & 0.496 & 0.664 & 10 \\
DRLT (ours) & 0.543 & 0.635 & 45 \\
DRLT-LSTM (ours) & 0.543 & 0.635 & 270 \\
\hline
\end{tabular}
\end{center}
\caption{AUC, precision scores and run-time speed comparison between ours and two state-of-the-art trackers. DRLT-LSTM is our tracker with pre-computed YOLO features and only computing LSTM transitions during tracking. Our tracker is tested on a single NVIDIA GTX 1080 GPU.}
\label{tb:comp_table}
\vspace{-3ex}
\end{table}

\begin{table*}
\scriptsize
\begin{center}
\begin{tabular}{|p{1.7cm}|K{1.1cm}|K{1.1cm}|K{1.1cm}|K{1.1cm}|K{1.1cm}|K{1.1cm}|K{1.1cm}|K{1.1cm}|K{1.1cm}|}
\hline
Sequence & DRLT & STRUCK & SCM & DLT & VTS & VTD & OAB & CSK & TLD\\
\hline\hline
Suv      & 0.621  & 0.519  & \textcolor{blue}{0.725} & \textcolor{red}{0.743} & 0.468 & 0.431 & 0.619 & 0.517 & 0.660  \\
Couple   & \textcolor{blue}{0.493}  & 0.484  & 0.100   & 0.237 & 0.057 & 0.068 & 0.346 & 0.074 & \textcolor{red}{0.761} \\
Diving   & \textcolor{red}{0.540}  & 0.235  & \textcolor{blue}{0.272} & 0.141 & 0.213 & 0.210  & 0.214 & 0.235 & 0.180  \\
Dudek    & 0.603   & 0.720   & 0.746 & 0.778 & \textcolor{red}{0.802} & \textcolor{blue}{0.789} & 0.653 & 0.707 & 0.643 \\
Human3   & \textcolor{red}{0.401}   & 0.007  & 0.006 & 0.007 & \textcolor{blue}{0.018} & 0.018 & 0.010  & 0.011 & 0.007 \\
Human6   & \textcolor{red}{0.413}  & 0.217  & 0.327 & \textcolor{blue}{0.342} & 0.168 & 0.168 & 0.207 & 0.208 & 0.282 \\
Human9   & \textcolor{red}{0.425}  & 0.065  & 0.138 & 0.165 & 0.111 & \textcolor{blue}{0.244} & 0.170  & 0.220  & 0.159 \\
Jump     & \textcolor{red}{0.452}  & \textcolor{blue}{0.105}  & 0.077 & 0.053 & 0.053 & 0.057 & 0.085 & 0.094 & 0.070  \\
Jumping  & 0.651 & \textcolor{red}{0.664}  & 0.116 & 0.598 & 0.149 & 0.116 & 0.069 & 0.050  & \textcolor{blue}{0.662} \\
Singer2  & \textcolor{red}{0.623} & 0.040   & 0.172 & 0.039 & 0.332 & \textcolor{blue}{0.416} & 0.045 & 0.043 & 0.026 \\
Skating1 & 0.457 & 0.285  & 0.444 & 0.422 & \textcolor{blue}{0.482} & \textcolor{red}{0.492} & 0.368 & 0.478 & 0.184 \\
Woman    & 0.479 & \textcolor{red}{0.693}  & \textcolor{blue}{0.651} & 0.595 & 0.132 & 0.145 & 0.466 & 0.194 & 0.129 \\
Dancer   & 0.685 & 0.625  & 0.708 & 0.571 & \textcolor{red}{0.728} & \textcolor{blue}{0.720}  & 0.604 & 0.609 & 0.394 \\
Liquor   & \textcolor{red}{0.532} & 0.408  & 0.311 & 0.342 & 0.427 & \textcolor{blue}{0.478} & 0.457 & 0.253 & 0.456 \\
BlurCar4 & 0.701 & \textcolor{red}{0.820}   & 0.416 & 0.657 & 0.079 & 0.079 & 0.777 & \textcolor{blue}{0.816} & 0.630  \\
Human7   & \textcolor{blue}{0.612} & 0.466  & 0.303 & 0.366 & 0.206 & 0.299 & 0.421 & 0.350  & \textcolor{red}{0.675} \\
Human8   & \textcolor{blue}{0.349} & 0.127  & \textcolor{red}{0.729} & 0.106 & 0.336 & 0.246 & 0.095 & 0.171 & 0.127 \\
BlurCar2 & 0.693 & \textcolor{red}{0.743}  & 0.185 & \textcolor{blue}{0.732} & 0.108 & 0.108 & 0.100   & 0.743 & 0.726 \\
Skater2  & \textcolor{red}{0.643} & 0.536  & 0.424 & 0.215 & 0.454 & 0.454 & 0.500   & \textcolor{blue}{0.546} & 0.263 \\
Bird2    & 0.473 & 0.565  & \textcolor{red}{0.756} & 0.221 & 0.328 & 0.214 & \textcolor{blue}{0.648} & 0.580  & 0.570  \\
Girl2    & \textcolor{red}{0.361} & 0.227  & \textcolor{blue}{0.266} & 0.058 & 0.257 & 0.257 & 0.071 & 0.060  & 0.070  \\
CarDark  & 0.548 & \textcolor{red}{0.872}  & \textcolor{blue}{0.844} & 0.582 & 0.717 & 0.521 & 0.765 & 0.744 & 0.423 \\
CarScale & 0.453 & 0.350   & \textcolor{red}{0.581} & \textcolor{blue}{0.539} & 0.436 & 0.442 & 0.325 & 0.400   & 0.434 \\
Car2     & 0.480  & 0.623  & \textcolor{blue}{0.908} & \textcolor{red}{0.909} & 0.774 & 0.774 & 0.569 & 0.652 & 0.660  \\
BlurCar3 & 0.680  & \textcolor{red}{0.780}   & 0.220  & 0.205 & 0.188 & 0.188 & \textcolor{blue}{0.720}  & 0.430  & 0.639 \\
Gym      & \textcolor{red}{0.641} & 0.350   & 0.214 & 0.178 & 0.359 & \textcolor{blue}{0.367} & 0.069 & 0.219 & 0.276 \\
BlurCar1 & 0.694 & \textcolor{blue}{0.760}   & 0.057 & 0.044 & 0.210  & 0.210  & \textcolor{red}{0.780}  & 0.011 & 0.605 \\
BlurBody & \textcolor{blue}{0.672} & \textcolor{red}{0.696}  & 0.194 & 0.145 & 0.238 & 0.238 & 0.671 & 0.381 & 0.391 \\
Skater   & \textcolor{red}{0.692} & 0.551  & 0.460  & \textcolor{blue}{0.556} & 0.470  & 0.471 & 0.481 & 0.431 & 0.326 \\
Dancer2  & \textcolor{red}{0.825} & \textcolor{blue}{0.776}  & 0.740  & 0.482 & 0.717 & 0.704 & 0.766 & 0.776 & 0.651 \\
\hline\hline
Average  & \textcolor{red}{0.562} & \textcolor{blue}{0.477}  & 0.403  & 0.368 & 0.334 & 0.331 & 0.402 & 0.367 & 0.403 \\
\hline
\end{tabular}
\end{center}
\caption{Average bounding box overlap ratio on individual sequence. \textcolor{red}{Red}: best, \textcolor{blue}{blue}: second best.}
\label{tb:seq_comp_table}
\end{table*}

Figure~\ref{fig:OPE_benchmark} illustrates the precision and success plots based on the center location error and bounding box overlap ratio, respectively. It clearly presented the superiority of our algorithm over other trackers. The higher success and precision scores indicated that our algorithm  hardly missed targets while maintaining good tracking of tight bounding boxes to targets. The superior performance was probably because the CNN captured representative features for localization and RNN was trained to force the long-term consistency of the tracking trajectory. 

To gain more insights about the proposed algorithm, we evaluated the performance of trackers on individual sequences in the benchmark. Table~\ref{tb:seq_comp_table} summarizes the average bounding box overlap ratio for each sequence. Our algorithm achieved best results for 12 sequences, and second best results for 4 sequences. We also achieved the best overall performance, beating the second best by almost 10\% (0.562 vs. 0.477).  Unlike other trackers where catastrophic failures were observed for certain sequences, our algorithm performed consistently well among all 30 sequences. This further illustrated that the spatial representations and temporal constraints learned by our algorithm were general, robust, and well-suited for tracking a large variety of targets. 

We compared our tracker qualitatively with two distinct benchmark methods as well as DLT in Figure~\ref{fig:qualitative}. It demonstrated that our tracker effectively handled different kinds of challenging situations that often required high-level semantic and temporal understanding such as motion blur, illumination variation, rotation, deformation, etc. Comparing with other trackers, our tracker hardly drifted to the background and  predicted more accurate and reasonable bounding box locations.

To verify the importance of RNN in our algorithm, we did more experiments on varying RNN step sizes. Step size denoted the number of frames considered each time for training the network, referred to as $T$ in Algorithm~\ref{alg:rltraining}. Success plots of three different RNN step sizes were illustrated in Figure~\ref{fig:internal}, and we found  that larger step sizes allowed us to model longer and more complicated temporal constraints, thus resulting in better accuracy. This analysis demonstrated the importance of incorporating temporal information in tracking and the effectiveness of using our RL formulation. 

Table~\ref{tb:comp_table} analyzed different trackers in terms of speed and accuracy. Our original model already operated at frame-rates beyond real-time by getting rid of online searching and fine-tuning mechanisms. Furthermore, pre-computing frame-level YOLO features off-line allowed us to only perform LSTM computation during online tracking, resulting in processing speed at 270 fps. In all, although our implementation was based on deep CNN and RNN, the proposed DRLT method was very efficient due to its extreme run-time simplicity, while preserving accurate tracking performance.

\section{Conclusion}
\label{sec:conclusion}

In this paper, we proposed a novel neural network tracking model based on a recurrent convolutional network trained with deep RL algorithm. To the best of our knowledge, we are the first to bring RL into CNN and RNN to solve the visual tracking problem. The entire network is end-to-end trainable off-line allowing it to run at frame-rates faster than real-time. The deep RL algorithm directly optimizes a long-term tracking performance measure which depends on the whole tracking video sequence. Other than CNN-based trackers, our paper aims to develop a new paradigm for solving the visual tracking problem by bringing in RNN and RL to explicitly exploit temporal correlation in videos. We achieved state-of-the-art performance on OTB public tracking benchmark.

We believed that our initial work shed light on many potential research possibilities along this direction. Not only better training and design of recurrent convolutional network can further boost the efficiency and accuracy for visual tracking, but a broad new way of solving vision problem with artificial neural network and RL can be further explored.  


{\small
\bibliographystyle{ieee}
\bibliography{egbib}
}

\end{document}